\documentclass[conference]{IEEEtran}
\IEEEoverridecommandlockouts
\usepackage{cite}
\usepackage{amsmath,amssymb,amsfonts}
\usepackage{algorithmic}
\usepackage{graphicx}
\usepackage{textcomp}

\usepackage{graphicx}   
\usepackage{adjustbox}  
\usepackage{multirow}   
\usepackage{booktabs}   

\usepackage{float}

\usepackage{xcolor}
\def\BibTeX{{\rm B\kern-.05em{\sc i\kern-.025em b}\kern-.08em
    T\kern-.1667em\lower.7ex\hbox{E}\kern-.125emX}}
\begin{document}

\title{Evaluating Robustness of Vision-Language Models Under Noisy
Conditions
\\
\thanks{Identify applicable funding agency here. If none, delete this.}
}

\author{\IEEEauthorblockN{1\textsuperscript{st} Purushoth Velayuthan}
\IEEEauthorblockA{\textit{Computer Science and Engineering} \\
\textit{University of Nevada Reno}\\
Reno, Nevada \\
pvelayuthan@unr.edu}
\and
\IEEEauthorblockN{2\textsuperscript{nd}Alireza Tavakkoli }
\IEEEauthorblockA{\textit{Computer Science and Engineering} \\
\textit{University of Nevada Reno}\\
Reno, Nevada  \\
tavakkol@unr.edu}
}

\maketitle

\begin{abstract}
Vision-Language Models (VLMs) have attained exceptional success across multimodal tasks such as image captioning and visual question answering. However, their robustness under noisy conditions remains unfamiliar. In this study, we present a comprehensive evaluation framework to evaluate the performance of several state-of-the-art VLMs under controlled perturbations, including lighting variation, motion blur, and compression artifacts. We used both lexical-based metrics (BLEU, METEOR, ROUGE, CIDEr) and neural-based similarity measures using sentence embeddings to quantify semantic alignment. Our experiments span diverse datasets, revealing key insights: (1) descriptiveness of ground-truth captions significantly influences model performance; (2) larger models like LLaVA excel in semantic understanding but do not universally outperform smaller models; and (3) certain noise types, such as JPEG compression and motion blur, dramatically degrade performance across models. Our findings highlight the nuanced trade-offs between model size, dataset characteristics, and noise resilience, offering a standardized benchmark for future robust multimodal learning.

\end{abstract}


\section{Introduction}
The arrival of Vision-Language Models (VLM) has revolutionized many sectors including au-
tonomous systems, robotics, healthcare, and real-world image understanding. VLMs incorporate
both textual and visual information, enabling the models to perform tasks such as scene descrip-
tion, image captioning, question answering and multimodal reasoning. Traditional vision-based
models mainly underperform with contextual ambiguity, creating urgency for data-hungry models
requiring large annotated datasets and handcrafted features to achieve robust performance. How-
ever, this issue has been partially addressed by VLMs that leverage self-supervised learning and
transformer-based architectures to enhance generalization across diverse real-world applications.
Despite the successes, VLMs still face challenges in overcoming complex environments with high
noise levels, occlusions and distortions. Images in the real world often contain variations in lighting,
movement blur and messy backgrounds, which affects the model’s robustness and reliability. In ad-
dition, tasks such as fine-grained visual understanding, spatial reasoning and long-term contextual
dependencies remain open challenges. Addressing these challenges, our research aims to evaluate
and develop strategies to enhance the robustness of VLMs under noisy conditions


To effectively evaluate the performance of VLM among diverse tasks and settings, we employ both lexical-based and neural-based evaluation methods. In the lexical category, we established metrics such as BLEU-1 to BLEU-4~\cite{papineni2002bleu}, which quantify the overlap between the generated and reference text based on n-gram precision. METEOR~\cite{banerjee2005meteor} complements this by ingesting synonym matching and stemming, offering a more significant assessment. ROUGE-L~\cite{lin2004rouge} evaluates the longest common subsequence between predicted and reference captions, effectively capturing fluency and sentence structure. CIDEr~\cite{vedantam2015cider} (Consensus-based Image Description Evaluation) places weight on rare yet informative n-grams to measure consensus with human-written captions, while SPICE~\cite{anderson2016spice} focuses on scene-graph similarity to evaluate semantic content.

In addition to these lexical metrics, we incorporate neural-based evaluation using the Sentence Transformer~\cite{koupaee2018wikihowlargescaletext} model. By encoding both the generated captions and ground truth descriptions into high-dimensional semantic vectors, we compute cosine similarity scores to assess how closely the VLM-generated outputs align semantically with the reference text. This allows us to capture refined meanings that purely lexical metrics might overlook, especially in paraphrased or restructured sentences.

Moreover, we evaluate the VLM's adaptability by comparing performance across older and newer image captioning datasets, using both baseline and state-of-the-art models. This comparative analysis gives light on how well models are able to generalize the different dataset characteristics, such as vocabulary diversity, image complexity, and descriptive richness. We further examine how the granularity and detail of the ground truth descriptions influence model performance.





The main contributions of this paper are as follows:

\begin{enumerate}
    \item We provide conclusive evidence on how the descriptiveness of ground truth captions significantly impacts the performance of VLMs  in caption generation.
    \item We demonstrate that state-of-the-art (SOTA) VLMs maintain strong performance in noisy environments, highlighting their robustness across varying levels of visual degradation.
    \item We identify and utilize an effective evaluation metric framework to comprehensively assess model performance across lexical and neural similarity metrics.
    \item We highlight that scaling up model size does not always improve performance on older datasets, as smaller models often align better with simpler, less descriptive captions.
\end{enumerate}

\section{Related Work}

Image captioning has long been a fundamental task in the vision-language paradigm. Initial approaches primarily integrated Convolutional Neural Networks (CNNs) for extracting visual features and Recurrent Neural Networks (RNNs) like LSTMs~\cite{chen2015mind} for generating captions sequentially. These early systems typically relied on disjoint visual and textual embedding spaces, which were later unified into a joint embedding space to better align image and language features. This evolution enabled models to generate captions that more accurately represented the core semantics of images. Attention mechanisms, Generative Adversarial Networks (GANs)~\cite{dai2017towards,shetty2017speaking}, and most recently, diffusion-based models~\cite{yu2025diffusion,jiang2024comat} have further refined the quality of caption generation.

Vision-Language Models (VLMs) such as CLIP, BLIP, BLIP-2,smolVLM LLava, PaliGemma etc. have since emerged as state-of-the-art solutions for multimodal tasks, including image captioning~\cite{luu2024questioning}, VQA~\cite{lee2023pix2struct}, and cross-modal retrieval~\cite{jiang2025visual}. Unlike traditional encoder-decoder models, VLMs offer a more unified architecture to handle multimodal reasoning tasks efficiently. However, while VLMs demonstrate remarkable performance in clean data settings, their robustness under real-world disturbances remains an open challenge. Captions generated by these models can degrade significantly in the presence of visual noise;such as occlusions, distortions, lighting variations, motion blur, compression artifacts, and background clutter;as well as textual noise like OCR errors, misspellings, or paraphrasing~\cite{kupyn2018deblurgan,yadav2023effective}.

To systematically evaluate these robustness concerns, Qiu et al. conducted large-scale evaluations on 12 vision-language models using 17 types of visual disturbances and 16 textual corruptions, proposing metrics such as Multimodal Impact Metrics (MMI) and Missing Object Rate (MOR) to measure degradation in caption quality~\cite{qiu2022multimodal}. Similarly, Park et al. introduced the ROCOCO benchmark with synthetically altered MS-COCO images, revealing substantial drops in recall metrics when evaluated on models like BLIP and VSE~\cite{park2023rococo}.

Various robustness-oriented training strategies have also been proposed. Rashid and Rivas explored fine-grained adversarial perturbations without full model retraining, while Zhou et al. introduced MMCOA~\cite{zhou2024revisiting}, a contrastive adversarial training method that aligns clean and perturbed multimodal embeddings. Other efforts like ALIP~\cite{yang2023alip} integrated synthetic captions and adaptive consistency gates to suppress noise during pretraining, further improving robustness. Notably, models such as TransferCLIP~\cite{cheng2024transfer} and RORA-VLM~\cite{qi2024rora} applied contrastive and adversarial strategies to strengthen noise tolerance, while BLIP-Diffusion and LLaVA-R utilized denoising and semantic cues to refine captioning under visual degradation.

Evaluating the captioning robustness of VLMs has required the development of specialized metrics. Traditional metrics such as BLEU, ROUGE, METEOR, and CIDEr provide n-gram or concept-based assessments. More recent embedding-based metrics like BERTScore~\cite{yi2020improving} and CLIPScore~\cite{chen2023research} account for semantic alignment under perturbation, showing greater resilience to lexical and paraphrastic noise. Advanced methods such as CLAIR leverage large language models like GPT, Claude, and PaLM for reference-free evaluation.

Despite these advancements, consistently assessing VLM performance under varied noisy conditions remains an open research area. Our work builds upon these insights to propose a unified evaluation framework designed to quantify both visual and textual robustness of VLM-based captioning systems using classical and embedding-based metrics under controlled perturbations, thereby ensuring fair and standardized comparisons across models.

\section{Methodology}

This section outlines the proposed method for comparing the performance of different VLM models on image-text datasets, with the goal of finding the model that is most robust to different levels of added noise in the images. 

\begin{figure}[H]
    \centering
    \includegraphics[width=0.5\textwidth]{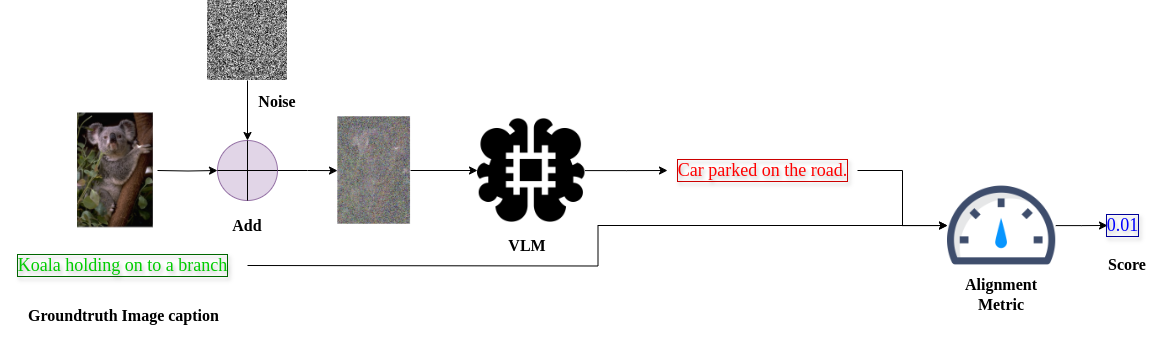}
    \caption{Overview of the evaluation pipeline: The VLM processes noisy images to generate captions, which are then compared against reference (golden) captions from the dataset using both lexical metrics (CIDEr, BLEU, METEOR, ROUGE) and neural metrics (CLIP Score, PR-MCS).}
    \label{fig:pipeline}
\end{figure}

Figure~\ref{fig:pipeline} shows the steps taken in process, namely dataset preparation, syntetic noise dataset creation, Model preparation, instruction prompt selection, evaluation which will be explained in depth in this section.

\subsection{Datasets}

To ensure a comprehensive evaluation across diverse visual scenes and linguistic contexts, we selected two prominent benchmarks: Flickr30k~\cite{plummer2016flickr30kentitiescollectingregiontophrase} and NoCaps~\cite{Agrawal_2019}. These datasets differ in terms of domain, vocabulary size, and scene complexity, offering a rigorous testbed for assessing the generalization ability of vision-language models.

The Flickr30k dataset, introduced in 2015, consists of 31,783 images collected from the Flickr photo-sharing platform. Each image is paired with five crowd-sourced captions describing the visual content. The dataset is rich in everyday scenes, human-object interactions, and casual settings, making it suitable for evaluating general-purpose captioning capabilities. For our experiments, we used the standard test split consisting of 1,000 images, ensuring consistency with prior works.

In contrast, the NoCaps dataset, published in 2019, was designed to test a model's performance in open-domain scenarios with a long-tail vocabulary. It extends the Open Images validation set with 15,100 images and over 166,000 reference captions, annotated with the help of human raters. Importantly, NoCaps includes three subsets in-domain, near-domain, and out-of-domain that vary based on object category overlap with COCO, allowing us to evaluate the models under both familiar and unfamiliar conditions. This enables a nuanced understanding of how noise affects both lexical recall and semantic grounding in zero-shot settings.

\subsection{Synthetic Noise Dataset Creation}
To rigorously evaluate the robustness of Vision-Language Models (VLMs) under diverse and challenging conditions, we will generate synthetic noisy variants of clean datasets (e.g. Flickr30k, Nocaps)
using a structured approach.

\subsection{Noise Levels and Mixtures}\label{AA}
We will first introduce per-noise-type degradations at incremental severity levels. For example, 
Gaussian noise will be applied with standard deviation ($\sigma$) values ranging from 0.1 (mild) to 1.0 (severe), while motion blur will vary in kernel size (5px–15px) and angle. 
Beyond isolated noise types, we will simulate real-world complexity by creating mixed noise conditions such as combining low-light adjustments (gamma correction, $\gamma \in [0.3, 1.5]$) with adversarial patches and sensor noise (Poisson/Gaussian, $\sigma \in [0.05, 0.3]$). 
These mixtures will test whether VLMs fail catastrophically when multiple distortions interact, as often occurs in practical deployments.

\subsection{Model Selection}
We conducted a detailed evaluation of Vision-Language Models (VLMs) under varying noise conditions, we selected a diverse set of state-of-the-art(SOTA) models: BLIP, BLIP-2, SmolVLM, SmolVLM-2, LLaVA-3B (Mistral), LLaVA-3B (Vicuna), PaliGemma, and PaliGemma-2. These models were chosen based on their architectural differences, pretraining strategies, and recent benchmark performance across image-captioning tasks. BLIP and BLIP-2 are well-known for their retrieval-augmented training and performs well on visual grounding tasks. SmolVLM variants are optimized for efficiency, and this makes them suitable for understanding performance trade-offs under constrained resources. LLaVA models ingest powerful LLMs (Vicuna and Mistral) with vision inputs, offering insights into how large-scale alignment impacts robustness. PaliGemma models which is delevoped by Google, are recent advancement in multilingual and multimodal grounding, aligning well with our interest in generalizable VLMs.

Each model was set up using publicly available checkpoints and repositories. For evaluation, we unified the pipeline across all models—resizing inputs, normalizing pixel values, and following model-specific tokenizer and vision encoder protocols. Caption predictions were generated for each image in both the clean and noisy datasets.

\subsection{Instruction Prompt selection}

 For the aim of evaluating caption generation across different Vision-Language Models (VLMs), we created a set of structured prompts. The template selection of prompts were selected based on the instruction-following capability of each model and aligned with standard zero-shot evaluation formats commonly used in VLM benchmarks. The prompts were grouped into three types according to their difficulty : (i) basic prompts (e.g., "Describe the image."), (ii) descriptive prompts (e.g., "List the objects and actions in the image."), and (iii) reasoning prompts (e.g., "What is happening in the image and why?").

Each prompt was adapted to the model’s input requirements. LLava and PaliGemma and their variants are basically were provided provided with system-style prompts that mirror their training format. Conversely, models such as BLIP and BLIP-2, which are not instruction-tuned, do not take in  directive-style prompts; instead, they produce output image-to-text generation using their default pretraining objective. For smolVLM and smolVLM-2, prompts were tailored to suit their lightweight architecture and instruction-tuned design.

To ensure uniformity across experiments, decoding parameters such as temperature, top-k sampling, and beam size were fixed throughout the evaluation process. This helped maintain consistency in caption generation and reduced variability due to randomness.

\subsection{Post-processing of Captions}

Every VLM model that we used follows a different decoding pipeline, often resulting in contrast in the format and structure of the generated captions. In particularly, models such as LLaVA and its variants (e.g., LLaVA-Vicuna, LLaVA-Mistral), as well as PaLI-Gemma and PaLI-Gemma2, prone to produce captions embedded within instruction-style responses. These outputs often include irrelevant character-level elements such as numbered lists, markdown formatting, hashtags, or newline characters: for example, captions beginning with sequences like \texttt{".\textbackslash n\textbackslash n3. **Banana Trees**: ..."} or containing symbols like \texttt{"\#"} and \texttt{"**"}.

To make sure consistency and comparability across models, we employed a post-processing strategy to clean and standardize the generated captions. This involved removing all non-alphanumeric characters (except for basic punctuation such as commas and periods), stripping markdown formatting, removing numbered or bulleted prefixes, and trimming leading/trailing whitespace. Regular expressions were used to identify and remove structured fragments (e.g., patterns like \#,**) commonly introduced by instruction-following models. This process ensured that only the core semantic content of the captions was retained, which is very important for fair and accurate evaluation using both lexical and semantic metrics.

\section{Evaluation}

\subsection{Evaluation Metrics}
We employ both lexical-based and nerual based evaluation methods to effectivly evaluate the performance of the VLM under different conditions and dataset variations. In the lexical methods, we used metrics such as BLEU-1 to BLEU-4, which compute the n-gram overlap between generated captions and ground-truth references. The METEOR metric boosts this by considering for stemming, synonymy, and word order, thereby capturing partial matches and improving alignment with human judgments. In case of ROUGE-L, focuses on the longest common subsequence to assess sentence structure and fluency in the predictions.

CIDEr, gives higher weights to rare but informative n-grams by using TF-IDF scores across multiple reference captions, aiding diversity and informativeness. 

In addition to these lexical-based metrics, we employ neural-based evaluation using Sentence-BERT, a Sentence Transformer model. Generated captions and ground-truth descriptions are embedded into dense vector spaces, and similarity between embeddings are caluculated using cosine similarity. This enables the quantification of semantic closeness beyond surface-level word matching, capturing subtle differences in meaning and structure, particularly in paraphrased or restructured captions.

\begin{table*}[htbp]
\centering
\caption{Evaluation Metrics Comparison across flickr\_1k and Nocaps Datasets}
\label{tab:big_ieee_table}
\begin{adjustbox}{max width=\textwidth}
\begin{tabular}{llcccccccccc}
\toprule
\textbf{Dataset} & \textbf{Model} & BLEU-1 & BLEU-2 & BLEU-3 & BLEU-4 & METEOR & ROUGE\_L & CIDEr & testlen & reflen & similarity \\
\midrule
\multirow{8}{*}{\textbf{flickr\_1k}} 
& Blip            & 0.6141 & 0.4028 & 0.2639 & 0.1704 & 0.1768 & 0.3652 & 0.3658 & 8705  & 9199  & 0.5819 \\
& Blip-2          & \textbf{0.6774} & 0.4640 & 0.3103 & 0.2057 & \textbf{0.2075} & \textbf{0.3994} & 0.4794 & 9371  & 9554  & 0.6492 \\
& Smol-vlm        & 0.4888 & 0.3289 & 0.2181 & 0.1438 & 0.1912 & 0.4511 & \textbf{0.6320} & 9923  & 9523  & 0.5909 \\
& Smol-vlm2       & 0.4852 & 0.3368 & 0.2278 & 0.1536 & 0.2592 & 0.3807 & 0.2297 & 21486 & 16495 & 0.6566 \\
& llava\_34b\_mist  & 0.5480 & 0.3841 & 0.2619 & 0.1768 & 0.2503 & 0.3934 & 0.3526 & 13750 & 12708 & \textbf{0.6727} \\
& llava\_34b\_vicuna& 0.5663 & 0.3978 & 0.2722 & 0.1835 & 0.2427 & 0.3902 & 0.3375 & 11749 & 11376 & 0.6708 \\
& PaliGemma       & 0.5886 & 0.4193 & 0.2914 & 0.1997 & 0.2264 & 0.3934 & 0.3438 & 9444  & 9546  & 0.6597 \\
& PaliGemma2      & 0.3455 & 0.1976 & 0.1149 & 0.0650 & 0.1225 & 0.2340 & 0.1374 & 7872  & 8859  & 0.5254 \\
\midrule
\multirow{8}{*}{\textbf{Nocaps}} 
& Blip            & 0.7000 & 0.5472 & 0.4055 & 0.2890 & \textbf{0.3466} & 0.4474 & 0.6351 & 37842 & 40937 & 0.7387 \\
& Blip-2          & 0.7404 & 0.5918 & 0.4487 & 0.3266 & 0.3324 & 0.4694 & 0.7273 & 35742 & 39737 & 0.7726 \\
& Smol-vlm        & 0.7138 & 0.5426 & 0.3977 & 0.2840 & 0.2346 & 0.4935 & 0.6125 & 50722 & 48787 & 0.7416 \\
& Smol-vlm2       & \textbf{0.7738} & 0.6185 & 0.4750 & 0.3558 & 0.2234 & 0.5546 & \textbf{0.7751} & 55529 & 53449 & 0.7690 \\
& llava\_34b\_mist  & 0.7664 & 0.6063 & 0.4602 & 0.3400 & 0.2567 & \textbf{0.5600} & 0.7359 & 55065 & 52555 & \textbf{0.7950} \\
& llava\_34b\_vicuna& 0.7658 & 0.6047 & 0.4571 & 0.3356 & 0.3456 & 0.5516 & 0.7127 & 49871 & 48889 & 0.7893 \\
& PaliGemma       & 0.7273 & 0.5718 & 0.4369 & 0.3265 & 0.3433 & 0.5282 & 0.7117 & 52096 & 50320 & 0.7339 \\
& PaliGemma2      & 0.4105 & 0.2419 & 0.1427 & 0.0829 & 0.2767 & 0.2915 & 0.2452 & 35035 & 40595 & 0.5933 \\
\bottomrule
\end{tabular}
\end{adjustbox}
\end{table*}

\begin{table*}[htbp]
\centering
\caption{Evaluation Metrics Comparison across flickr\_1k with Noisy Generated Dataset}
\label{tab:big_ieee_table2}
\begin{adjustbox}{max width=\textwidth}
\begin{tabular}{lllcccccccccc}
\toprule
\textbf{Noise Type} & \textbf{Level} & \textbf{Model} & Ratio & BLEU-1 & BLEU-2 & BLEU-3 & BLEU-4 & METEOR & ROUGE-L & CIDEr & Similarity \\
\midrule
\multirow{3}{*}{Adversarial} 
& High   & Blip-2       & 0.889 & 0.5700 & 0.3548 & 0.2175 & 0.1306 & 0.1478 & 0.3340 & 0.2523 & 0.3911 \\
& Low    & Blip-base    & 0.970 & 0.5296 & 0.3123 & 0.1837 & 0.1087 & 0.1309 & 0.3043 & 0.2085 & 0.5113 \\
& Medium & Llava-vicuna & 0.989 & 0.4539 & 0.2958 & 0.1815 & 0.1107 & 0.1730 & 0.2966 & 0.1585 & 0.4390 \\
\midrule
\multirow{3}{*}{Defocus Blur} 
& High   & Llava-vicuna & 0.979 & 0.4461 & 0.2339 & 0.1126 & 0.0595 & 0.0915 & 0.2564 & 0.0632 & 0.2388 \\
& Low    & Paligemma    & 1.010 & 0.3553 & 0.1901 & 0.0952 & 0.0480 & 0.1081 & 0.2219 & 0.0489 & 0.4943 \\
& Medium & Smolvlm2     & 0.966 & 0.3254 & 0.1752 & 0.0768 & 0.0352 & 0.0950 & 0.2011 & 0.0221 & 0.3407 \\
\midrule
\multirow{3}{*}{Gaussian Blur} 
& High   & Smolvlm      & 0.972 & 0.2836 & 0.1431 & 0.0677 & 0.0317 & 0.0652 & 0.1741 & 0.0145 & 0.2616 \\
& Low    & Paligemma2   & 0.986 & 0.3075 & 0.1732 & 0.0928 & 0.0497 & 0.1015 & 0.2025 & 0.0310 & 0.3336 \\
& Medium & Blip-base    & 0.978 & 0.2498 & 0.1207 & 0.0550 & 0.0250 & 0.0702 & 0.1684 & 0.0081 & 0.2507 \\
\midrule
\multirow{3}{*}{Gaussian Noise} 
& High   & Blip-2       & 0.992 & 0.4813 & 0.2959 & 0.1697 & 0.0991 & 0.1252 & 0.2842 & 0.1371 & 0.4710 \\
& Low    & Smolvlm      & 1.004 & 0.3795 & 0.2136 & 0.1069 & 0.0535 & 0.0917 & 0.2364 & 0.0617 & 0.4312 \\
& Medium & Smolvlm2     & 1.018 & 0.4572 & 0.2880 & 0.1683 & 0.0986 & 0.1634 & 0.3034 & 0.1359 & 0.4979 \\
\midrule
\multirow{3}{*}{Impulse Noise} 
& High   & Llava-mistral& 0.986 & 0.3961 & 0.2419 & 0.1393 & 0.0800 & 0.1214 & 0.2531 & 0.0804 & 0.4181 \\
& Low    & Blip-base    & 0.996 & 0.5125 & 0.3124 & 0.1895 & 0.1127 & 0.1467 & 0.3160 & 0.2112 & 0.4822 \\
& Medium & Blip-2       & 0.992 & 0.4623 & 0.2815 & 0.1625 & 0.0935 & 0.1297 & 0.2870 & 0.1342 & 0.4490 \\
\midrule
\multirow{3}{*}{Motion Blur} 
& High   & Smolvlm2     & 0.994 & 0.3311 & 0.1764 & 0.0885 & 0.0467 & 0.0918 & 0.2132 & 0.0302 & 0.3620 \\
& Low    & Paligemma    & 1.002 & 0.4057 & 0.2338 & 0.1223 & 0.0678 & 0.1157 & 0.2578 & 0.0839 & 0.4533 \\
& Medium & Blip-base    & 0.990 & 0.3679 & 0.2028 & 0.1058 & 0.0580 & 0.1024 & 0.2359 & 0.0602 & 0.4112 \\
\midrule
\multirow{3}{*}{Snow} 
& High   & Blip-2       & 0.986 & 0.4295 & 0.2560 & 0.1437 & 0.0846 & 0.1182 & 0.2705 & 0.1015 & 0.4372 \\
& Low    & Smolvlm      & 0.993 & 0.3857 & 0.2156 & 0.1152 & 0.0619 & 0.0957 & 0.2447 & 0.0693 & 0.3921 \\
& Medium & Llava-vicuna & 0.981 & 0.4092 & 0.2367 & 0.1310 & 0.0742 & 0.1275 & 0.2620 & 0.0867 & 0.4216 \\
\midrule
\multirow{3}{*}{Zoom Blur} 
& High   & Paligemma2   & 0.984 & 0.3014 & 0.1651 & 0.0864 & 0.0448 & 0.0894 & 0.2061 & 0.0279 & 0.3295 \\
& Low    & Smolvlm2     & 0.987 & 0.3620 & 0.1975 & 0.1018 & 0.0542 & 0.0973 & 0.2257 & 0.0470 & 0.3849 \\
& Medium & Llava-mistral& 0.978 & 0.3368 & 0.1845 & 0.0926 & 0.0491 & 0.0910 & 0.2150 & 0.0354 & 0.3578 \\
\bottomrule
\end{tabular}
\end{adjustbox}
\end{table*}

\subsection{Results}

Table~\ref{tab:big_ieee_table} shows the evaluation results of selected VLMs on two datasets: the Flickr validation set (Flickr\_1k) and the Nocaps dataset. Out of the two, the Nocaps dataset resulted higher lexical scores across all metrics—BLEU-1 (0.7738), METEOR (0.3466), ROUGE\_L (0.5600), and CIDEr (0.7751)—as well as the highest neural similarity score (0.7950), showing that its richer, more descriptive annotations gives a stronger caption quality assessment .

The results also shows that no single model outperforms others across all metrics. On the Flickr dataset, the smaller model BLIP2 gives a result of  the best BLEU-1 score (0.6774), while on Nocaps, smolVLM2 performs best in terms of BLEU-1 (0.7738). In contrast, the larger model LLaVA recieves the highest similarity scores for both datasets (0.6727 for Flickr and 0.7950 for Nocaps), mentioning its strength in capturing semantic alignment rather than lexical overlap.

Table~\ref{tab:big_ieee_table2} compiles the performance of the selected models under various noise types and levels. Notably, the PaliGemma2 model achieved the highest similarity score under pixelation noise (0.6324), outperforming all other models. In contrast, the most hardest noise condition was JPEG compression at a high noise level, where models struggled significantly (lowest score: 0.2280). Addition to this, motion blur under high noise conditions capitulate the overall lowest similarity scores, further confirming the impact of noise type and intensity on model robustness.

\section{Discussions}

\begin{figure}[H]
  \centering
  \includegraphics[width=0.9\linewidth]{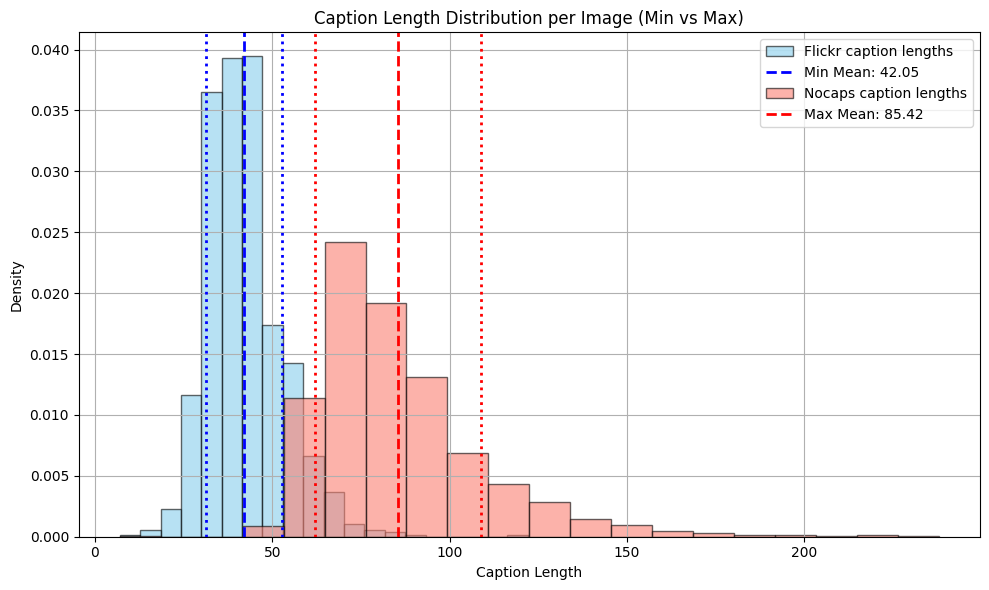}
  \caption{Overview of Flickr caption length and Nocaps caption length }
  \label{fig:example}
\end{figure}

This research mainly targets on evaluating the robustness of VLMs) in noisy environments. We experimented to examine how well these models handle noise and maintain performance across different degradation levels. As a initial point of baseline experiment, we first evaluated the models on clean datasets to understand how ground truth captions influence the quality of the generated outputs.

While it is more intuitively assumed that larger, state-of-the-art (SOTA) models perform better on standard descriptive image datasets, our experiments points out otherwise in certain instances. As shown in Table~\ref{tab:big_ieee_table}, smaller models often outperform larger ones in aligning generated captions with ground truth, particularly on older datasets such as Flickr. These datasets tend to feature more common and less descriptive captions, which may align better with the output style of smaller models.

Contrastively, as described in Table~\ref{tab:caption_comparison}, the Nocaps dataset—being more recent—contains richer and more descriptive annotations. SOTA models, designed to generate detailed and semantically more dense captions, reuslts to perform better on such datasets. This is further supported by the caption length distribution in Figure~\ref{fig:example}, which shows that modern image-text datasets consist of longer, more descriptive sentences than older ones like Flickr. Table~\ref{tab:caption_examples} gives example captions from both Nocaps and Flickr,points the difference in detail of the captions. Captions from Nocaps often include the main object and its actions, while Flickr captions are more general. This is evidented out by the similarity scores in Table~\ref{tab:big_ieee_table}, where SOTA models shows a strong alignment with ground truth on richly annotated datasets.

Table~\ref{tab:big_ieee_table2} further shows that many SOTA models holds strong performance in noisy environments. This robustness can be assigned to their training on large-scale datasets encircling diverse conditions, along with their huge parameter density that allow them to capture fine-grained variations. The hardest and type of noise also play a significant role: when the noise level is low or minimally distorts the visual content (e.g., Speckle Noise, Gaussian Blur), model performance does not drop considerably. However, more heavy noise conditions makes a notable performance degradation.

A comparison between Table~\ref{tab:big_ieee_table} and Table~\ref{tab:big_ieee_table2} shows a key insight: models that outputs high performance on clean datasets and do not necessarily maintain the same performance when evaluated on noise-induced versions of those datasets. This gives out the importance of evaluating VLMs not just under ideal conditions but also in scenarios that reflect real-world imperfections and distortions.





\section*{Conclusion}

This work evaluates the robustness of Vision-Language Models (VLMs) under varying noise levels and dataset characteristics. Our findings show that smaller models can outperform larger ones on older datasets with simpler captions, challenging the assumption that bigger models always perform better. In contrast, SOTA models excel on newer, richly annotated datasets like Nocaps.

We also show that many SOTA models maintain strong performance under mild noise, but face degradation under heavy distortions. These results highlight the importance of evaluating VLMs not only on clean datasets but also in noisy, real-world scenarios to ensure practical reliability.



\begin{table}[!t]
\caption{Comparison between descriptive (Nocaps) and less descriptive (Flickr) image captions}
\label{tab:caption_comparison}
\centering
\small 
\begin{tabular}{|p{3.7cm}|p{3.7cm}|}
\hline
\textbf{Descriptive Captions (Nocaps)} & \textbf{Less Descriptive Captions (Flickr)} \\
\hline
A baby is standing in front of a house & Dogs racing \\
\hline
A little girl in a white jacket and sandals running around in the garden & Dogs playing football \\
\hline
A young child stands in front of a house & A man in a bar going outside \\
\hline
\end{tabular}
\end{table}





\bibliographystyle{ieeetr}

\bibliography{conference_101719}
\end{document}